# Autonomous 3D Reconstruction Using a MAV

Alexander Popov, Dimitrios Zermas and Nikolaos Papanikolopoulos

*Abstract –An approach is proposed for high resolution 3D reconstruction of an object using a Micro Air Vehicle (MAV). A system is described which autonomously captures images and performs a dense 3D reconstruction via structure from motion with no prior knowledge of the environment. Only the MAV's own sensors, the front facing camera and the Inertial Measurement Unit (IMU) are utilized. Precision agriculture is considered as an example application for the system.*

## INTRODUCTION

Most modern high definition 3D scans are done using RGB-D cameras. In recent years, with the introduction of the Microsoft Kinect, 3D pointclouds became a hot topic of research [1]. A multitude of projects have been under development, such as those from the Microsoft KinectFusion and MIT [2], [3]. 3D point clouds are also at the core of many robotics applications, ranging from Simultaneous Localization and Mapping (SLAM) and artificial intelligence to augmented reality [4], [5]. However, common RGB-D sensors, such as the Microsoft Kinect, are heavy, require a large energy supply and do not work in outdoor environments. In particular, they use infrared sensors to estimate depth and thus sunlight heavily interferes with their measurements. Therefore, another method is needed to acquire the missing depth measurements outdoors and on small vehicles.

As an alternative, we propose to use small, low cost cameras such as those found on the Parrot AR Drone quadrocopter. Small cameras have become widespread throughout industry and household devices. They are cheap and are a great source of information. Therefore our system uses a single moving camera for extracting the missing depth information. The Parrot AR Drone quadrocopter, given its low cost and sensor rich hardware, proved to be an excellent platform and a great example of a low cost MAV with a front facing camera [6]. Our system uses the AR Drone to combine a multitude of different techniques for extracting the 3D information in order to perform SLAM and a dense 3D reconstruction from just a single camera.

Bundle Adjustment and Structure-from-Motion are used to perform dense reconstruction given a set of images with multiple views of the same scene. This has been demonstrated in systems such as the Bundler, the Clustering Views for Multi-view Stereo (CMVS) and the Patch-based Multi-View Stereo (PMVS), [11]–[13]. While the combination of these structure from motion tools provides a great alternative to RGB-D cameras, they come at a very heavy computational cost. Bundle Adjustment and the following CMVS and PMVS processes are extremely computationally demanding. Therefore, using them in real-time for SLAM and navigation is impractical and a new approach must be found.

To address this problem other methods have been developed such as PTAM (Parallel Tracking and Mapping) and the MonoSLAM software packages [14], [15]. Ultimately, combining these techniques, our system utilizes sparse and dense reconstruction to demonstrate a method for obtaining a dense point cloud from a monocular system while localizing and navigating in real-time. We use PTAM to create real-time sparse 3D reconstructions which allow us to perform SLAM for navigation [16]. We further utilize DBSCAN to cluster the sparse map and to identify the target object in near real-time[17]. We then plan a path and circle the object to capture the required images for dense 3D reconstruction, which is done in parallel, using the tools mentioned above. One of our contributions is the use of clustering in real-time to identify objects in a sparse 3D map for object detection, navigation and dense 3D reconstruction. By separating sparse and dense reconstruction we obtain a system which can navigate a MAV in real time using a mono-slam camera (sparse reconstruction) and IMU readings while in parallel performing dense 3D reconstruction for later use.

We believe that applications using low cost MAVs for detailed 3D reconstructions can be the key in enabling a multitude of application such as the scenario of precision agriculture. We currently know of no other system which with such low cost can provide detailed 3D scans in outdoor environments.

While quite a few works have been published,[7]–[10], describing how a MAV can be used to preform large area field scans for precision agriculture. Our work is aimed at extending these ideas by showing how a MAV can be utilized to perform detailed high resolution scans for precision agriculture.

## THE SYSTEM AND 3D RECONSTRUCTION

Related work can be broken into three parts: tracking via a monocular system, navigating a MAV using a single camera and IMU readings, and, dense 3D Reconstruction via Structure from Motion.

Tracking and localizing a single camera in an unknown environment is a very challenging problem. The main issue arises with the inability to easily extract depth from a single image. Thus, the only alternative is to either have a stereo system or to perform structure-from-motion reconstruction. However, in our case, doing the latter in real time poses very high computational demands and until recently it was an open research problem [18]. One of the major breakthroughs came with the publication of PTAM (Parallel Tracking and Mapping) by Georg Klein [14]. The breakthrough of the PTAM is the realization that the two key components of structure from motion can be parallelized and executed on two CPUs at once. As most modern computers have at least two CPU cores, it seemed like a reasonable assumption. The first task, tracking key points from frame to frame in order to solve the correspondence problem, and to localize the camera given the reconstructed points can be executed on one CPU; while the second task, solving the multiple view geometry and mapping the key points as fast as the process can keep up, can be done on another. The mapping portion is done using the Bundle Adjustment [19]. Another benefit of using the PTAM is the fact that the software is openly published under a research license, eliminating the need to re-invent the wheel.

It is important to notice that due to the great computational constraint of extracting depth for even just a few hundred points, PTAM performs only a sparse reconstruction by selecting maximal FAST corners - this poses some problems [20]. Without real-time dense map, navigation and localization becomes more difficult, not to mention the need for other tools to perform the dense reconstruction later on.

At this point it is also important to discuss the pitfalls of the PTAM. Specifically, there are issues that arise due to the sparsity of the map. Tracking can fail when there is too much rapid motion as the few tracked corners (key points) can be lost. Further, the system assumes a static scene. If there is too much motion or the scene changes drastically, the system will fail as it will no longer find the required key points. Finally, since the system relies on corner detection, it is assumed that the scene has enough of them to build the required map.

Navigating the MAV is another challenge. There are a couple of other works which discuss using the Kalman Filter to integrate the IMU readings for MAV control [21]–[23]. We would like to focus on the work from the Technical University of Munich (TUM) which has used the Extended Kalman Filter to better estimate the position of the Parrot AR Drone quadrocopter by fusing its on board IMU readings and the pose estimation generated by cameras using PTAM [16]. They have also released the associated software compatible with the ROS (Robot Operating System) [24]. Their work provides a good starting point for navigating a low cost quadrocopter, such as the AR Drone, using its own cheap IMU sensors and the front facing camera. ROS provides the necessary platform to merge all of these tools together.

The Bundler, the CMVS and the PMVS packages, first published at the University of Washington, provide an offline solution for creating dense 3D point clouds given a set of unordered overlapping images [11]–[13]. The above projects focused on a large set of unordered images, such as those from readily available tourist photo albums, using which they create 3D reconstructions of tourist destinations. We utilize the same tools and apply them to the images from the quadrocopter to generate dense 3D reconstruction of target objects. Since this set of tools is just one of the available options for performing dense reconstruction offline, and since this portion is completely interchangeable in our system, we will not go into further details describing the particular tools. However, the reader is strongly encouraged to become familiar with them for a complete understanding of our work. In summary, these tools take a large set of unordered images, compute the correspondences between the overlapping portions in those images and then compute sparse, followed by a dense reconstruction of the image set, generating a dense 3D point cloud.

### A. An Overview of the system

Our research involved modifying the existing tools to create a system which can autonomously guide a MAV to identify an object and then to create a dense 3D reconstruction of that object (Figure 1). We have adapted and modified the TUM_Ardrone ROS package, released by the TUM to extract the PTAM's sparse 3D map of the environment [27]. We have further created a ROS node which in real-time filters and clusters the sparse map using the DBSCAN in order to determine the target object which we want to scan for dense reconstruction [17]. In addition, we have created a PID controller for the AR Drone in order to guide it to an identified object, which it then circles in order to capture images at specified waypoints for dense reconstruction. These images are used to generate dense 3D reconstruction in parallel by using the Bundler, the CMVS and the PMVS software packages.

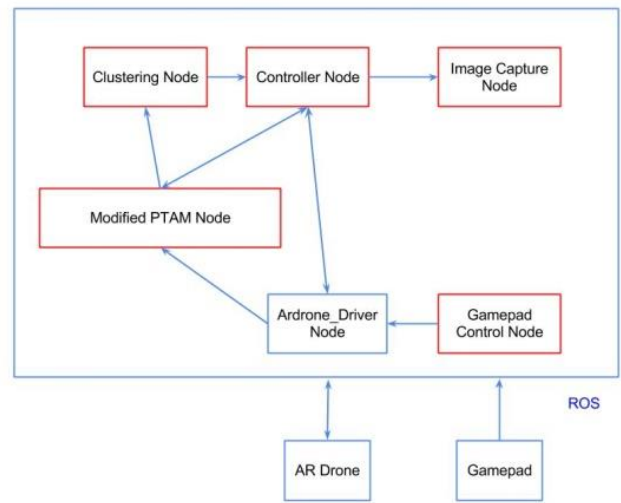

Figure 1 System overview.

### B. Map Extraction

To obtain the sparse map of the environment we have modified the TUM_Ardrone ROS package. Our modifications extract the current sparse 3D map as soon as it is generated in real time by the PTAM. The map is then published in ROS as a standard PCL Pointcloud2 sparse map.

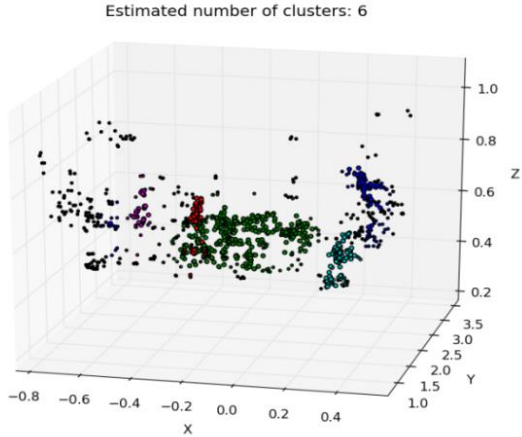

Figure 2  Full, unfiltered, sparse map of a scene (red points identify the closest object to the camera).

### C. Clustering

Once the map is published, the clustering ROS node consumes it for further processing (Figure 2). We filter the map based on the *y* coordinate, keeping the closest 10% of values to the drone (Figure 4). This threshold was determined empirically. Once filtering is done, we use the DBSCAN algorithm to cluster the points, empirically choosing *Eps* to be .99 and minimum cluster size to be 20. These values seemed to provide the best results given the average density of the PTAM map and assuming that the object we are looking for has a significant number of key features. These values can be changed for other scenes or more intelligent detection algorithms.

More formally, the e*ps-neighborhood* of a point *p* can be defined by the following equation [17]:

$$N_{Eps}(p) = q_i \in D \mid dist(p,q) \leq Eps \quad (1)$$

where *D* is the database of points in our 3D space at a particular 3D reconstruction frame, and $q_i$, where *i* goes from 1 to the number of points in *D*, represents all other points in *D*. In our case *eps*, the distance between points, equals to .99, as described above.

Further, our point *p* must also meet the following conditions in order to be identified as being in a cluster *C*, which we believe to be the object which we are looking for:

1. $NumPoints\left(N_{Eps}(p)\right) > MinPoints$ (2)
2. $p \in 10^{th}$ percentile of points closest to the camera (3)

In our case, *MinPoints* was empirically determined to be 20 as described above. It must also be mentioned that a point *p* must be "density-connected" to all points *q* with respect to *eps* and MinPoints in order to be in the cluster. However, this goes outside the scope of our paper and is heavily elaborated on in [17]. At the end, we pick a cluster *C*, where all of the points in the cluster meet the above criteria.

The DBSCAN algorithm proved to be efficient at filtering out noise and selecting, *in real-time*, prominent objects in the map.

Once the clustering is complete, the node then selects the closest cluster to the MAV and publishes its world coordinates on a ROS topic. While other, more sophisticated, approaches can be used for selecting an appropriate object, this initial approach proved adequate as a proof of concept.

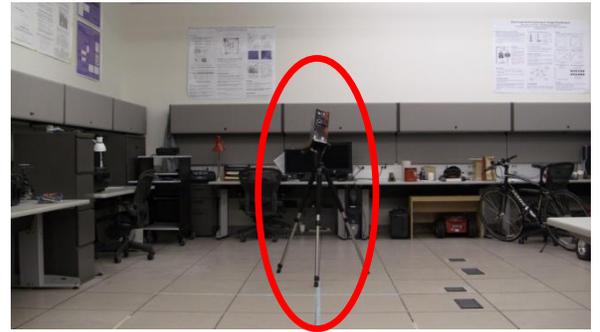

Figure 3  Tripod.

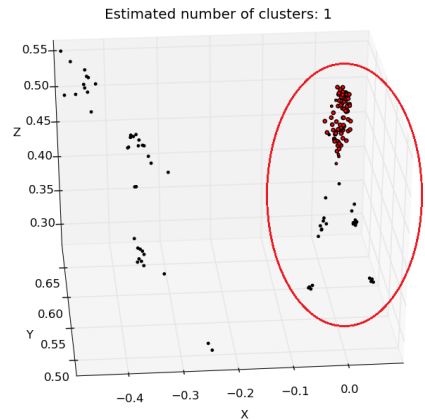

Figure 4  Filtered and clustered sparse reconstruction of the tripod.

### D. Controller

The controller ROS node receives the pose estimation from the modified *PTAM ROS node* and the location of the target object cluster from the *clustering ROS node*. As a demo controller, it plans a circular path around the detected object using the process described in the previous section (Figure 6). The reason we chose a circular path is to ensure we capture all of the views of the target in order to provide sufficient overlapping images to the PMVS/CMVS dense reconstruction software. A PID controller is used to plan and follow a circular path with waypoints at selected degrees on the circumference around the object (Figure 5). We attempt to keep the yaw rotation speeds low in order to try to maintain PTAM's key point tracking for as long as possible; PTAM struggles to maintain tracking through rotation around the z-axis (yaw) due to motion blur and triangulation errors in the sparse 3D reconstruction which arise due to the lack of translation (Figure 7). If the AR Drone loses visual tracking, the position is maintained by the Extended Kalman Filter alone as described above and in [16]. We start by selecting the starting point as our current position right in front of the object. We then map a circle with the object always being in the center, planning enough waypoints to ensure that we cover all of the angles of the object. The number was chosen empirically based on the starting distance away from the object. Whenever, the drone reaches a waypoint around the circle, the controller node publishes a message on a ROS topic which the separate *image capture ROS node* can then use to save the current image.

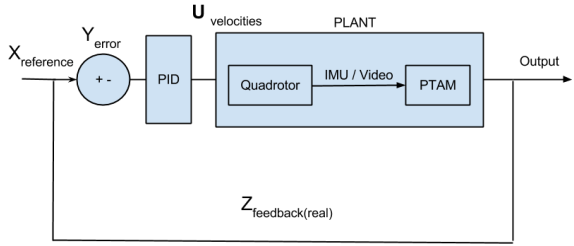

Figure 5 PID controller overview.

Given the quick execution of the PID controller the coefficients $K_I$ and $K_S$ are ignored, as on average, no large spikes or oscillations were observed and in turn a simpler model was chosen for the controller (Equations (4) through (8)).

$$X_{reference} = \begin{bmatrix} x \\ y \\ z \\ yaw \end{bmatrix} * R_z(\theta) \quad (4)$$

$$Z_{feedback} = \begin{bmatrix} x \\ y \\ z \\ yaw \end{bmatrix} \quad (5)$$

$$U_{velocity} = (X_{reference} - Z_{feedback}) * PID \quad (6)$$

$$PID = K_p + \frac{K_I}{S} + K_D * S \quad (7)$$

$$K_p = \begin{bmatrix} K_X \\ K_Y \\ K_Z \\ K_{yaw} \end{bmatrix}. \quad (8)$$

It is also important to mention that the waypoints are created and stored in the initial world coordinate frame. Therefore, the x and y coordinates of the next target waypoint must be multiplied by the rotation matrix $R_z(\theta) = \begin{pmatrix} \cos\theta & \sin\theta \\ -\sin\theta & \cos\theta \end{pmatrix}$, where $\theta$ indicating the amount of degrees the drone has rotated on the z-axis away from its original reference frame, before they can be used by the controller.

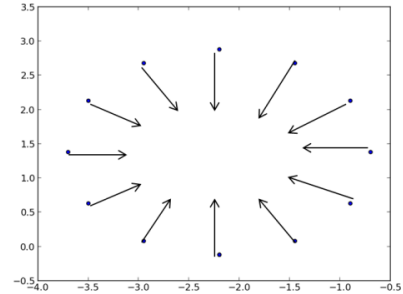

Figure 6 Reference coordinates the drone will follow. The center of the circle is at the target object.

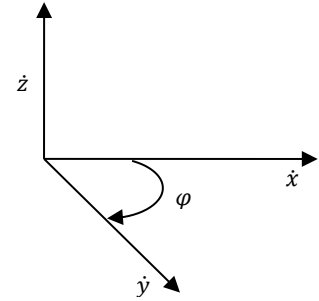

Figure 7 AR Drone degrees of freedom as exposed by the driver.

### E. Image Capture

The *image capture ROS node* waits for a "waypoint reached" message published by the controller. While it waits and to ensure that the correct frame is captured, it buffers the latest frame in memory; the node acquires frames from the ardrone_driver node in ROS. Whenever, the "waypoint acquired" message is received, the node saves the latest image. After at least two waypoints are reached, the *image capture ROS node* can start the dense 3D reconstruction.

### F. Dense Reconstruction

To do the dense reconstruction we use the Bundler, the CMVS, and the PMVS software packages. The process first determines the overlapping images, creates a sparse reconstruction, and then a dense reconstruction of the scene. This portion is done as a separate process as it can take a significant amount of time. Other tools may be used to do this, but the ease of use of the above tools made them a perfect candidate for this application. After these tools complete the reconstruction, the user is given a dense 3D reconstruction .ply file of the scene which contains the target object in the center. We have left these tools unmodified from their original versions, thus a more detailed description of the process can be found in the associated papers [11]–[13]. A sample result can be seen in Figure 8. Further dense reconstructions using a high resolution camera which could be mounted on the drone are shown in Figures 9 and 12.

It is also important to point out, that due to the large lens distortion on the AR Drone, a lens distortion correction filter must be applied on the images before the above procedures can commence.

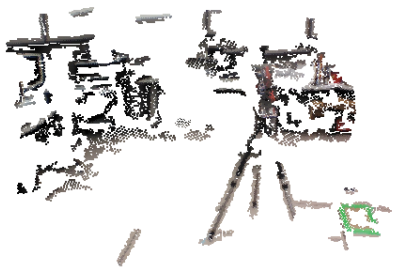

Figure 8 Sample dense reconstruction.

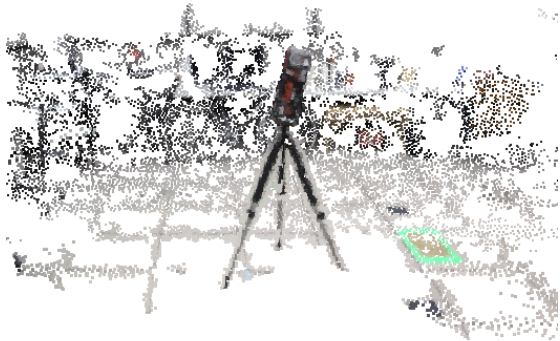

Figure 9 Dense reconstruction using a high resolution camera.

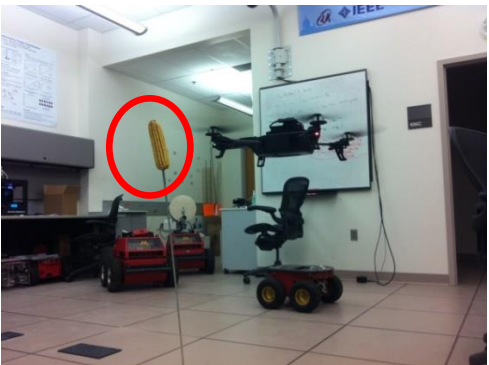

Figure 10 AR Drone flying in front of a corn cob.

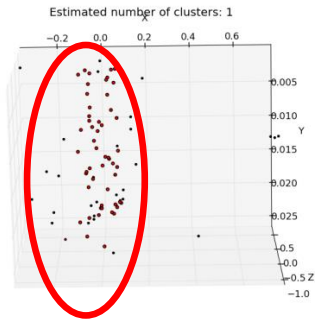

Figure 11 Sparse reconstruction of a corn cob; it is being identified as the closest object.

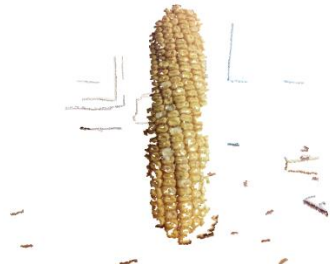

Figure 12. Dense reconstruction of a corn cob using a high resolution camera.

## CONCLUSION

The work was motivated by the prospective application of autonomous high resolution 3D scanning in hard to reach places. In particular we analyzed the approach in the context of precision agriculture. A method and the currently available tools are described for autonomously obtaining a dense 3D reconstruction given a monocular camera on a low cost MAV. The MAV first localizes itself in the environment using a sparse map, detects the target object using real-time clustering, plans and follows a path around the object for image capture and then performs dense 3D reconstruction using the Bundler, the CMVS, and the PMVS software packages. The whole system is incorporated into ROS for easier distribution and to make it easily compatible with future projects.

There is a lot of potential for future work in this area. Of course, the next natural step is to perform live dense reconstruction. There has been some significant progress in this area with works such as the DTAM, and it does seem that this should be feasible soon [25], [26].

Another direction for further study should be associated with cloud robotics. With platforms such as Roboearth, the computationally expensive need of having to constantly rebuild the map and the model of the object could be eliminated, and the computational resources could be directed towards other tasks, such as searching for the key object in the environment [28]. The system could have a database of healthy and unhealthy crops where a team of drones could scan large areas and zone in on problem spots for detailed scans.

An ability to have a low cost quadrocopter fly out and sample key portions of corn plants to identify Nitrogen deficient areas is an example where our system could be utilized [29]. Having a dense 3D reconstruction can provide the farmer with a detailed view of the health of the corn. The 3D data gives the ability to analyze the size of and shape of the corn on all of its sides.

In this work, sample results in a lab environment are demonstrated as proof-of-concept-principles for using the developed system in such a scenario. Particularly, the idea of expanding the use of an MAV from large area analysis in agriculture to high resolution scans of particular objects is explored and a possible solution for addressing this problem using a low cost MAV is presented.

## ACKNOWLEDGEMENTS

The authors thank Dr. Duc Fehr for his inputs and technical expertise. This material is based upon work supported by the National Science Foundation through grants #IIP-0934327, #IIP-1032018, #IIS-1017344, #CNS-1061489, #CNS-1138020, #IIP-1127938, #IIP-1237259, and #IIP-1332133.